%% file: main.tex
\documentclass[10pt,twocolumn,letterpaper]{article}

\usepackage{cvpr}              


\input{defs}

\definecolor{cvprblue}{rgb}{0.21,0.49,0.74}
\usepackage[pagebackref,breaklinks,colorlinks,allcolors=cvprblue]{hyperref}

\def\paperID{7} 
\def\confName{UNCV-W}
\def\confYear{2025}

\title{View-Dependent Uncertainty Estimation of 3D Gaussian Splatting}

\author{Chenyu Han \qquad Corentin Dumery\\
Computer Vision Lab, EPFL\\
Lausanne, Switzerland\\
{\tt\small chenyu.han@epfl.ch} \qquad {\tt\small corentin.dumery@epfl.ch}
}

\begin{document}
\maketitle
\input{sec/0_abstract}    
\input{sec/1_intro}
\input{sec/2_related}

\input{sec/3_method}
\input{sec/4_experiments}
\input{sec/5_conclusion}
{
    \small
    \bibliographystyle{ieeenat_fullname}
    \bibliography{main}
}


\end{document}

%% file: defs.tex
\newif\ifdraft
\draftfalse

\definecolor{burntorange}{rgb}{0.8, 0.33, 0.0}
\definecolor{orange}{rgb}{1,0.5,0}
\definecolor{green0}{rgb}{0.1,0.7,0.1}

\ifdraft
\newcommand{\PF}[1]{{\color{red}{\bf PF: #1}}}

\newcommand{\CD}[1]{{\color{blue}{\bf CD: #1}}}

\newcommand{\CH}[1]{{\color{green0}{\bf CH: #1}}}

\else
\newcommand{\PF}[1]{{\color{red}{}}}

\newcommand{\CD}[1]{{\color{blue}{}}}

\newcommand{\CH}[1]{{\color{green0}{}}}

\fi


%% file: sec/0_abstract.tex
\begin{abstract}
3D Gaussian Splatting (3DGS) has become increasingly popular in 3D scene reconstruction for its high visual accuracy. However,  uncertainty estimation of 3DGS scenes remains underexplored and is crucial to downstream tasks such as asset extraction and scene completion. 
Since the appearance of 3D gaussians is view-dependent, the color of a gaussian can thus be certain from an angle and uncertain from another.
We thus propose to model uncertainty in 3DGS as an additional view-dependent per-gaussian feature that can be modeled with spherical harmonics. This simple yet effective modeling is easily interpretable and can be integrated into the traditional 3DGS pipeline. It is also significantly faster than ensemble methods while maintaining high accuracy, as demonstrated in our experiments.
\end{abstract}

%% file: sec/1_intro.tex
\section{Introduction}
\label{sec:intro}
3D reconstruction is a fundamental task in computer vision, with applications in a wide range of areas such as video games, special effects or autonomous driving. The development of neural networks has brought new ideas to the field of 3D reconstruction, including neural radiance fields (NeRF), which use neural networks to implicitly represent 3D scenes. Even more recently, 3D Gaussian Splatting (3DGS)~\cite{kerbl3Dgaussians} proposed to use Gaussian particles to represent scenes, resulting in better performance while allowing real-time rendering.\\
Uncertainty estimation is important in 3D reconstruction, with various downstream applications. For example, in object completion, it is highly desirable to have a view-dependent uncertainty metric as it distinguishes between well-defined and previously unseen points of views. Similarly, in active view selection, an agent needs to find the viewpoint that reduces uncertainty in the scene the most. Many approaches have been proposed for uncertainty estimation of NeRF models, including naive methods like ensemble, and Bayesian Inference methods which are directly or indirectly related to model parameters~\cite{goli2023bayesraysuncertaintyquantification}. However, research on uncertainty estimation of 3DGS scenes is relatively limited, and most existing works apply ideas borrowed from NeRF uncertainty estimation to 3DGS~\cite{jiang2024fisherrfactiveviewselection}. However, the explicit representation of 3DGS brings significant differences and calls for an explicit modeling of uncertainty, which would be inherently interpretable and easily integrated in the 3DGS pipeline. 

In this paper, we propose a new uncertainty estimation method, which is tailored to 3DGS and takes full advantage of its capabilities. We draw inspiration from the view-dependent color representation of Gaussians, and propose to model uncertainty as an additional feature of Gaussians. This is simply achieved by representing the uncertainty using spherical harmonics to model their view dependency, similar to the view-dependent color. This additional learnable per-gaussian uncertainty value can easily be integrated into the existing 3DGS training pipeline. Experiments show that our method yields results similar to baseline methods like ensemble, with a significant advantage in performance and interpretability. 

After reviewing previous works on 3D reconstruction and uncertainty quantification in \cref{sec:rw}, we formally define 3DGS uncertainty and propose a simple yet effective view-dependent modeling in~\cref{sec:method}. Finally, we evaluate its performance and interpretability in~\cref{sec:exp}.

%% file: sec/2_related.tex
\section{Related Work}\label{sec:rw}
\label{sec:formatting}

\subsection{3D Reconstruction}
In early works, traditional methods including multiview stereo (MVS) and Bundle Adjustment laid the foundation for 3D scene reconstruction. Seitz \textit{et al.}~\cite{seitz2006comparison} provided a comprehensive evaluation of various MVS algorithms. Following the advent of neural networks, neural radiance fields (NeRF) have been proposed by \cite{mildenhall2020nerf}, which uses a network to model a continuous volumetric scene function to synthesize novel photorealistic views from sparse input images. More recently, 3D Gaussian Splatting (3DGS) \cite{kerbl3Dgaussians} has proposed to represent the scene with a set of 3D Gaussians, achieving high rendering speeds as well as high rendering quality. Its explicit representation of the reconstructed scene also brings significant advantages in terms of interpretability and editability.

\subsection{Uncertainty Estimation in 3D Reconstruction}
Methods for uncertainty estimation have advanced along with the reconstruction methods themselves. \cite{morris1999uncertainty}, \cite{szeliski1997shape} and \cite{wilson2020visual} have been proposed to estimate uncertainty using traditional methods. After NeRF was introduced, there have been many works on estimating the uncertainty for NeRF~\cite{goli2023bayesraysuncertaintyquantification,sunderhauf2023density,shen2021stochastic}. Some of them target irreducible uncertainties that come from motion blurs, non-static scene, or camera parameters. Other works focus on reducible uncertainties resulting from missing data due to occlusions, ambiguities, or limited camera views. Sünderhauf et al. \cite{sunderhauf2023density} proposed an adaption of the ensemble method, which is very costly in time and resources as it requires optimizing several reconstructions from different random initializations. Shen et al. \cite{shen2021stochastic} designed an additional Bayesian neural network based on the original NeRF to measure the uncertainty. Goli et al. \cite{goli2023bayesraysuncertaintyquantification} proposed the \textit{Bayes' Rays} method, which trains a perturbation field to estimate uncertainty. 

Few works have focused on uncertainty estimation in the specific case of 3DGS. Jiang \textit{et al.}~\cite{jiang2024fisherrfactiveviewselection} proposed a method for estimating the uncertainty of Gaussian parameters in 3DGS, and tried to extend the approach to the pixel level. Wilson \textit{et al.}~\cite{wilson2024modelinguncertainty3dgaussian} described a method that calculates the contribution of different camera input views to each Gaussian. Many previous works have focused on uncertainties of model parameters, and have no consideration of the view-dependency of Gaussians \cite{jiang2024fisherrfactiveviewselection}. These works typically consider the uncertainty of Gaussian as the average uncertainties of its parameters. For downstream applications, such as object completion, it is highly desirable to have a view-dependent uncertainty metric as it distinguishes between well-defined and previously unseen points of views. Similarly, in active view selection, an agent needs to find the viewpoint that reduces uncertainty in the scene the most. In that case, a view-independent uncertainty metric may not be able to correctly guide the agent to under-observed positions. This is the main drawback of previous methods. In contrast, a view-dependent uncertainty can provide better estimation of the scene uncertainty for such applications.

%% file: sec/3_method.tex
\section{Method}\label{sec:method}

\subsection{Preliminaries}
\paragraph{3DGS rendering.}
3D Gaussian Splatting uses Gaussians to represent scenes. Each Gaussian has an anisotropic color $c$, opacity $\alpha$, position $p$, and a 3D covariance matrix $\Sigma$ decided by rotation $R$ and scale $S$. The rendering process of 3DGS computes the color of each pixel by alpha-blending alone the pixel ray $x$. In this case, the rendered color of pixel $x$ can be represented as: 
$$C(x)=\sum_{i=1}^{N} c_n(x) T_i k_i$$
with
$$k_i = \alpha_i\texttt{exp}(-\frac{1}{2}(x-p)^{T}\Sigma^{-1}(x-p)), T_i = \prod_{j=1}^{i-1}(1-k_j)$$
Let $C_{gt}(x)$ be the ground truth color value of pixel $x$. If pixel $x$ is part of the training set, in a well-trained 3DGS model, we have
$$C_{gt}(x) \approx C(x)$$

\paragraph{Sources of uncertainty.} Klasson \textit{et al.} \cite{klasson2024sources} identify 4 types of uncertainties in 3D reconstruction. In this work, we mainly focus on \textit{reducible} uncertainty, which \textit{comes from insufficient information, and can be reduced by capturing data from new poses.} Other sources of uncertainty, including capture noise, inconsistencies, or motion blur, are not considered. We assume that the captured images are accurate, the camera poses are correctly estimated, and the scene is static.
\paragraph{Types of uncertainty.} Previous works have distinguished color uncertainty from position uncertainty~\cite{goli2023bayesraysuncertaintyquantification}. We argue that in 3DGS, both types of uncertainty can be considered part of \textit{Gaussian} uncertainty that denotes the uncertainty of the Gaussians themselves. 
\paragraph{Spherical harmonics.} In 3DGS, each Gaussian has an anisotropic color that can be modeled as a color function $c:\mathbb{S}^2\rightarrow~R^3$, where the color of a given Gaussian is a function of the view direction. This function is usually implemented using spherical harmonics (SH). Spherical harmonics are widely used to approximate spherical functions such as view-dependent lighting. SH are a set of orthonormal basis functions defined on the surface of a sphere, and thus an arbitrary spherical function can be decomposed into a weighted sum of basis functions. To address the view-dependency, we also use spherical harmonics to represent the uncertainty.

\subsection{Problem Statement}
\begin{figure}[htbp]
    \begin{minipage}[c]{0.5\linewidth}
        \centering
        \includegraphics[height=\textwidth]{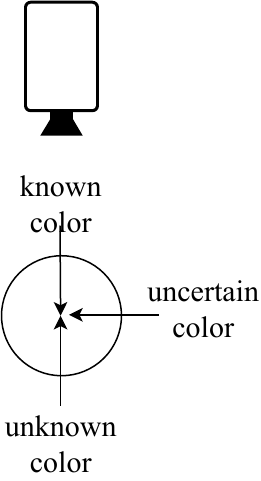}
        \centerline{(a) Color uncertainty}
    \end{minipage}%
    \begin{minipage}[c]{0.5\linewidth}
        \centering
        \includegraphics[height=\textwidth]{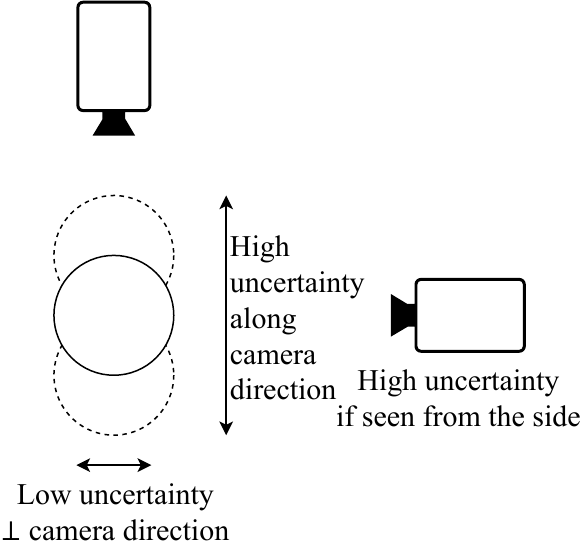}
        \centerline{(b) Position uncertainty}
    \end{minipage}
    \caption{Uncertainty types intuition}
    \label{fig:intuition}
\end{figure}

In the previous section, we have limited our definition of uncertainty to reducible uncertainty. This type of uncertainty stems from insufficient input data and is therefore highly related to the positions of training cameras. Intuitively, the visible parts of Gaussians should be considered as certain, and the invisible parts should be considered as uncertain, as shown in~\cref{fig:intuition}. In this figure, we also intuitively show why both types of uncertainty can be considered part of Gaussian uncertainty. Moreover, this figure also indicates that, the uncertainty of a Gaussian may vary in different view directions, and thus the uncertainty should be view-dependent. This view-dependency can then be modeled by an additional field per-gaussian expressed with spherical harmonics.

\subsection{Uncertainty Quantification}
\subsubsection{Definition}
As discussed in the previous section, the uncertainty value should be anisotropic since we are only confident of the color value from a certain pixel-ray direction. From other directions, the color value should be uncertain, and the direction that is far different from the pixel ray direction has a higher uncertainty. We thus represent uncertainty with spherical harmonics, similarly to the color representation of Gaussians.

For clarity, let us assume a simple case where a single Gaussian contributes to the color of a given pixel, and after optimization we have
$$
C_{gt}(x) \approx C(x) \approx c_0(x)
$$
where $C(x)$ is the rendered color, and $c_0(x)$ the color of the Gaussian particle. As we observe the rendering result to be close from the ground truth, we are confident that there is no uncertainty in this pixel, and thus no uncertainty in the Gaussian from the pixel-ray direction. Let $u$ be the proposed uncertainty of the Gaussian, then we aim to optimize this new attribute such that $u_0(x) = 0$ as seen from direction $x$.

To this end, we model the uncertainty as a function $u:\mathbb{S}\rightarrow (0, 1)$. Intuitively, each input camera provides some \textit{certainty} to the Gaussians that are visible from that angle. In this sense, our definition of uncertainty should be directly related to its visibility. A Gaussian visible from more cameras, or directly visible without being occluded, is of more certainty. 
We thus propose to project Gaussians onto training cameras, and supervise all Gaussians that contribute to a pixel by at least a certain threshold $T$. This makes it straightforward to supervize their uncertainty $u$, as explained in the following section.

\subsubsection{Training and Implementation}
We aim to decrease uncertainty from training views while maintaining high values from unseen angles. This naturally leads to the following loss:
$$\mathcal{L}_u =(1-\lambda)(u(x)-0)+\lambda(\bar{u}-1)$$
where $u(x)$ is the uncertainty of a given Gaussian,  and $\bar{u}$ denotes the average uncertainty value across directions. In practice, $\bar{u}$ is estimated via random sampling. In this formulation, the first term ensures low uncertainty in known directions, and the second term can be seen as a form of regularization keeping uncertainty values high. However, we observed that computing $\bar{u}$ is needlessly costly and instead propose to use the following expression:
$$\mathcal{L}_u =(1-\lambda)(u(x)-0)+\lambda(u(-x)-1)$$
where $u(x)$ denotes the uncertainty value of the Gaussian from pixel ray $x$ direction, and $u(-x)$ denotes the uncertainty value from the opposite direction of the pixel ray $x$. The hyperparameter $\lambda$ should be strictly smaller than $0.5$, which ensures that if a Gaussian is visible from two opposite directions, the uncertainty is still low from both viewpoints. 

Since the spherical harmonics are inherently smooth, this second loss formulation still works satisfyingly. According to our experiments, the results obtained by these two loss functions are similar. 
These two versions of uncertainty do not consider the alpha-blending process. Ideally, we would compute the \textit{contribution} of each Gaussian to a pixel, to produce more accurate results. However, this simplified approach is easy to implement in the traditional 3DGS pipeline, and in the following section we will show that it is sufficient to achieve satisfactory results in the majority of cases. 

%% file: sec/4_experiments.tex
\section{Experiments}\label{sec:exp}
\subsection{Dataset and Metrics}
We evaluated our uncertainty estimation method with the nerfstudio~\cite{nerfstudio} \textit{poster} dataset, which contains one static scene, consists of 226 consecutive images extracted from a video. We selected 50 random and consecutive images as the evaluation sets. The purpose of selecting consecutive images is to construct an uncertain angle.
We first visualize the view-dependency of our uncertainty estimation by rendering from different directions within the same scene in~\cref{fig:view_dependency}. Then we qualitatively evaluate the estimation by comparing with the baseline \textit{ensemble} method, and use the Area Under Sparsification Error (AUSE) with Mean Absolute Error (MAE) to quantitatively evaluate the estimation. To compute the AUSE metric, we select some of the input images as the evaluation set, and gradually remove pixels from the image, first in uncertainty order (\textit{i.e.} removing the most uncertain pixels first), and then in error order. The difference between the MAE of these two removal processes, or sparsification, is evaluated. The lower the difference, the better the uncertainty estimate.
The ensemble result is based on 10 copies of trained models with completely random initializations, which is computationally very heavy. The implementation of our method is based on the \textit{splatfacto} method from the \textit{nerfstudio} library, and the uncertainty value is obtained during the regular training process, with little additional performance consumption.
\subsection{Results}
\begin{figure}[htbp]
    \centering
    \begin{minipage}[t]{0.24\linewidth}
        \centering
        \includegraphics[width=\textwidth]{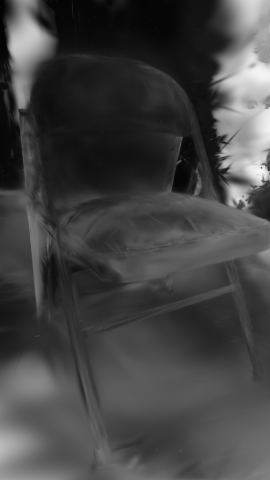}
    \end{minipage}%
    \hfill
    \begin{minipage}[t]{0.24\linewidth}
        \centering
        \includegraphics[width=\textwidth]{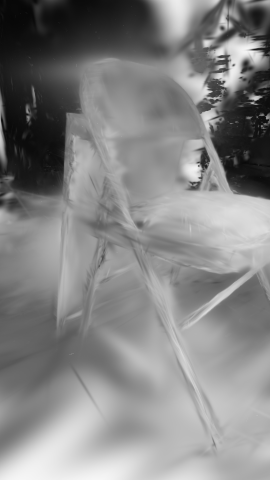}
    \end{minipage}%
    \hfill
    \begin{minipage}[t]{0.24\linewidth}
        \centering
        \includegraphics[width=\textwidth]{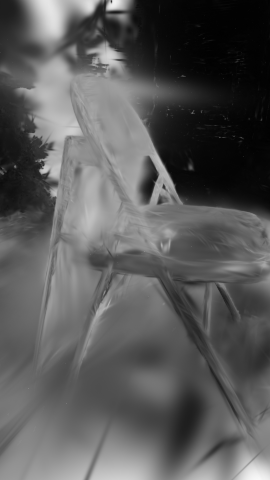}
    \end{minipage}%
    \hfill
    \begin{minipage}[t]{0.24\linewidth}
        \centering
        \includegraphics[width=\textwidth]{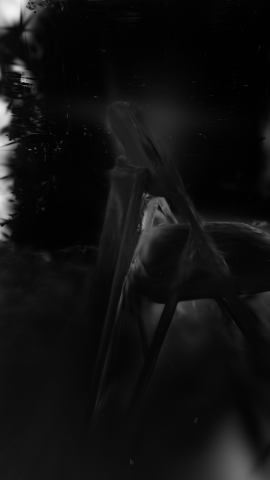}
    \end{minipage}%
    \caption{Uncertainty visualization from different angles, some of which are close to the training views.}
    \label{fig:view_dependency}
\end{figure}
\begin{figure}[htbp]
    \centering
    \begin{minipage}[t]{0.32\linewidth}
        \centering
        \includegraphics[width=\textwidth]{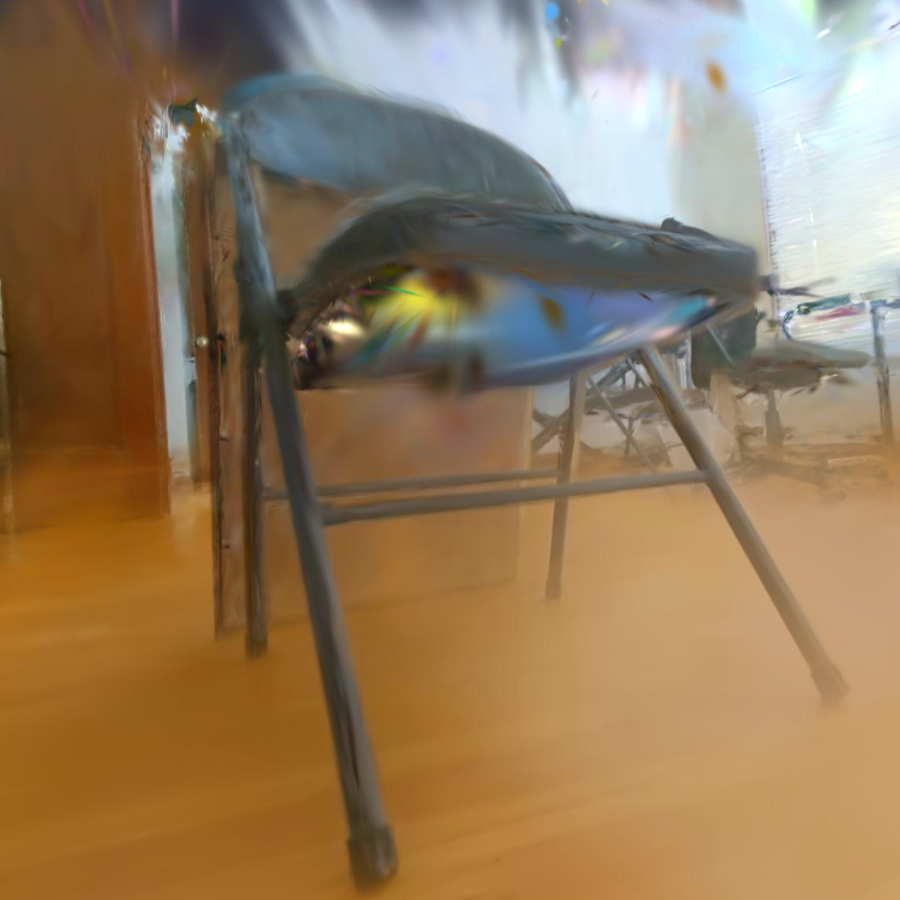}
        \centerline{(a) Render}
    \end{minipage}%
    \hfill
    \begin{minipage}[t]{0.32\linewidth}
        \centering
        \includegraphics[width=\textwidth]{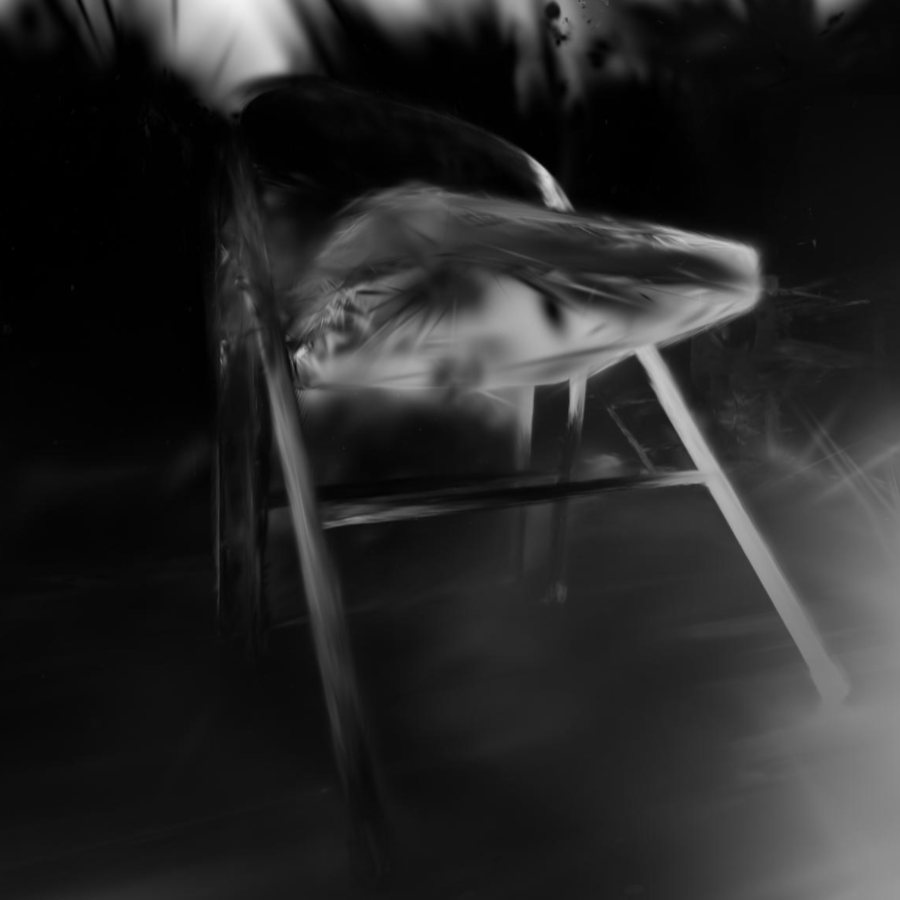}
        \centerline{(b) Ours}
    \end{minipage}%
    \hfill
    \begin{minipage}[t]{0.32\linewidth}
        \centering
        \includegraphics[width=\textwidth]{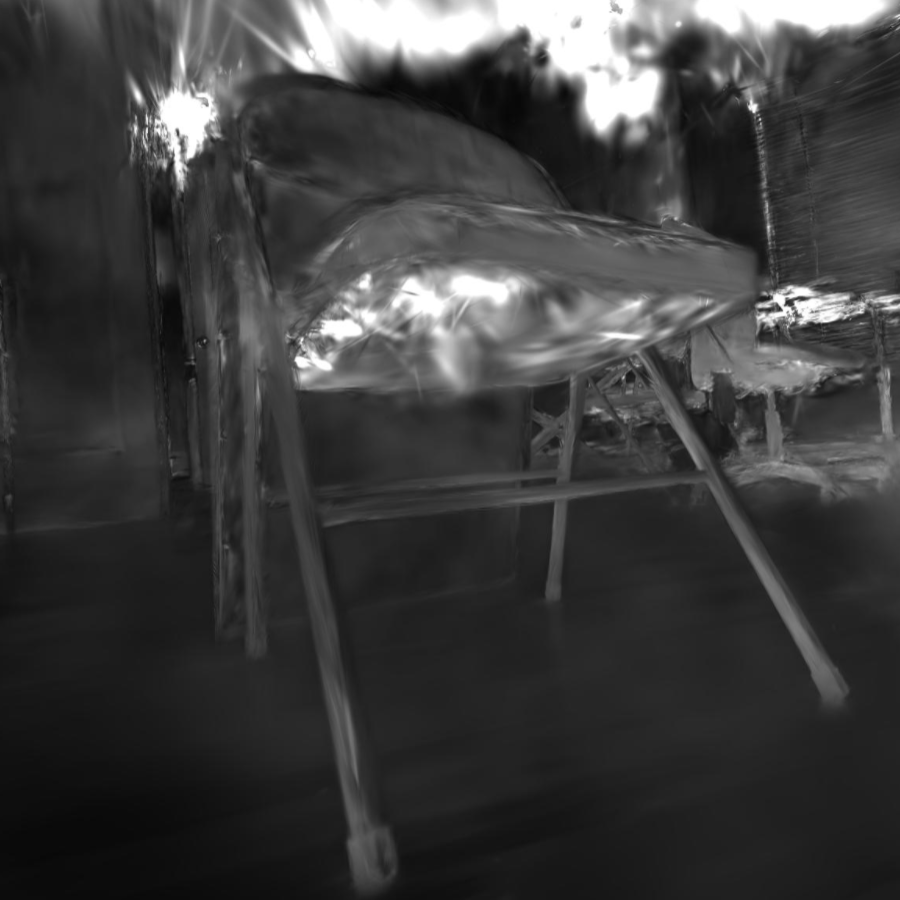}
        \centerline{(c) Ensemble}
    \end{minipage}%
    \caption{Uncertainty visualization from an unseen view in the \textit{poster} scene.}
    \label{fig:qualitative}
\end{figure}

\cref{fig:view_dependency} shows the uncertainty estimation from different view directions of a certain scene. The first and last images are from directions that are close to input images, and the middle images are relatively far from the input image directions. We can see that our method produce different results for different view directions.

We also compare our approach again the \textit{ensemble} method. \cref{fig:qualitative} shows the estimation result qualitatively in the \textit{poster} dataset. This scene contains a chair and a poster leaning against the chair. The input only contains images from above and sides; thus intuitively, the uncertainty of the bottom of the chair should be high. Our method correctly displays the high uncertainty of the bottom and is close to the result of the baseline ensemble.
\begin{figure}[htbp]
    \centering
    \begin{minipage}[c]{\linewidth}
        \centering
        \includegraphics[width=\textwidth]{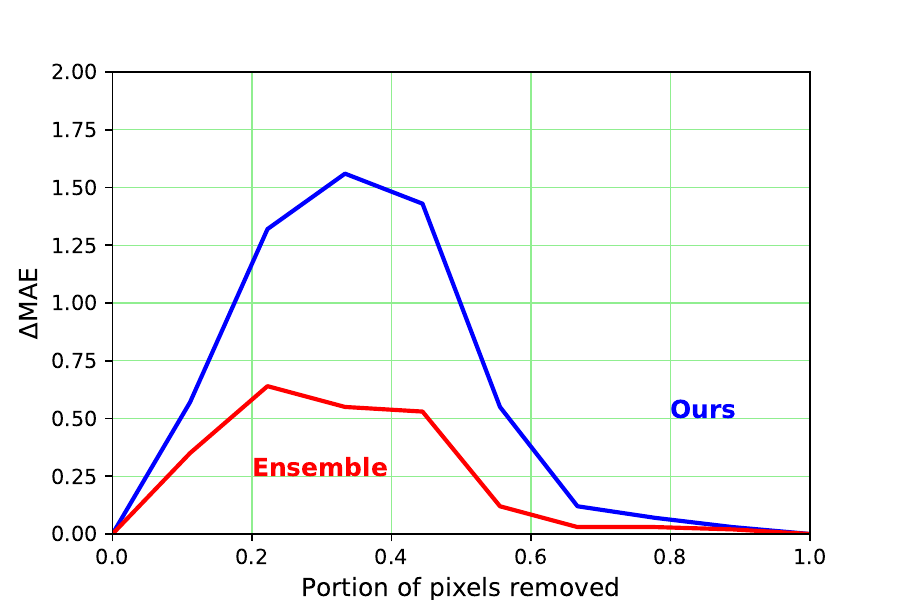}
    \end{minipage}
    \caption{A sparsification of the \textit{poster} scene.}
\end{figure}

\begin{table}[htbp]
    \centering
    \begin{tabular}{c|cc|c}
    \hline
    & \textit{poster} & \textit{chocolate} & Time (min)\\
    \hline
    Ours & 0.57 & 0.41 & 20\\
    Ensemble ($10\times$) & 0.23 & 0.11 & 169\\
    \hline
    \end{tabular}
    \caption{AUSE evaluation result.}
    \label{tab:quantitative}
\end{table}

\cref{tab:quantitative} shows the AUSE evaluation result.

Although our method does not outperform \textit{ensemble}, it still produces reasonable results and is not far behind. The reason for this discrepancy result is mainly due to our simplification, as explained in the previous section. In \cref{fig:qualitative}, we can see that our method can produce a wrong estimation at the edges of objects.

%% file: sec/5_conclusion.tex
\section{Conclusion}

We introduced a novel method for view-dependent uncertainty estimation of 3DGS. Our method is explicit and interpretable by design, unlike existing works. It can be easily implemented and integrated into existing training pipelines, with only a slight drop in accuracy compared with the expensive ensemble method. \\
In the future, we will expand our proposed uncertainty to take the alpha-blending weights into account in its formulation, and demonstrate the advantages of a view-dependent uncertainty for applications such next view selection or object completion.